\documentclass{article}

\usepackage{PRIMEarxiv}

\usepackage[utf8]{inputenc} 
\usepackage[T1]{fontenc}    
\usepackage{hyperref}       
\usepackage{url}            
\usepackage{booktabs}       
\usepackage{amsfonts}       
\usepackage{nicefrac}       
\usepackage{microtype}      
\usepackage{lipsum}
\usepackage{fancyhdr}       
\usepackage{graphicx}       
\graphicspath{{media/}}     
\usepackage{caption}
\usepackage{csquotes}
\usepackage{doi}
\usepackage{longtable}
\usepackage{caption}
 
\usepackage{graphicx} 
  
\title{Combining Knowledge Graphs and Large Language Models}

\author{
  Amanda Kau\thanks{These authors contributed equally to this work.}\\
  Australian National University \\
  Canberra, Australia\\
  \texttt{amanda.kau@anu.edu.au} \\
   \And
  Xuzeng He\footnotemark[1] \\
  Australian National University \\
  Canberra, Australia\\
  \texttt{u7189309@anu.edu.au} \\
   \And
  Aishwarya Nambissan \\
  Australian National University \\
  Canberra, Australia\\
  \texttt{Aishwarya.Nambissan@anu.edu.au} \\
  \And
  Aland Astudillo \\
  Swinburne University of Technology \\
  Victoria, Australia\\
   \texttt{aastudillocontreras@swin.edu.au} \\
   \And
Hui Yin \\
 Swinburne University of Technology \\
 Victoria, Australia\\
  \texttt{ huiyin@swin.edu.au} \\
   \And
 Amir Aryani \\
 Swinburne University of Technology \\
 Victoria, Australia\\
  \texttt{ aaryani@swin.edu.au} \\
}

\begin{document}
\maketitle

\begin{abstract}
In recent years, Natural Language Processing (NLP) has played a significant role in various Artificial Intelligence (AI) applications such as chatbots, text generation, and language translation. The emergence of large language models (LLMs) has greatly improved the performance of these applications, showing astonishing results in language understanding and generation. However, they still show some disadvantages, such as hallucinations and lack of domain-specific knowledge, that affect their performance in real-world tasks. 
These issues can be effectively mitigated by incorporating knowledge graphs (KGs), which organise information in structured formats that capture relationships between entities in a versatile and interpretable fashion. Likewise, the construction and validation of KGs present challenges that LLMs can help resolve. The complementary relationship between LLMs and KGs has led to a trend that combines these technologies to achieve trustworthy results. This work collected 28 papers outlining methods for KG-powered LLMs, LLM-based KGs, and LLM-KG hybrid approaches.
We systematically analysed and compared these approaches to provide a comprehensive overview highlighting key trends, innovative techniques, and common challenges. This synthesis will benefit researchers new to the field and those seeking to deepen their understanding of how KGs and LLMs can be effectively combined to enhance AI applications capabilities.
\end{abstract}


\section{Introduction}\label{sec_Introduction}
The rapid advancement of natural language processing (NLP) in recent years can be attributed to the availability of large datasets and the surge in computing power. 
Consequently, numerous large language models (LLMs) have been developed, such as Google's BERT \cite{devlin2019bert} and T5~\cite{raffel2023exploringlimitstransferlearning}, OpenAI's GPT series\cite{radford2018improving}.
LLMs are widely used in various tasks, including language translation, content creation, and virtual assistants. They excel in text generation, enabling applications such as automated essay writing, report generation, and creative storytelling. LLMs provide highly accurate translations in language translation, facilitating communication across different languages. They are also used in chatbots and customer service to handle queries efficiently and provide personalized responses. Additionally, LLMs assist in summarizing large volumes of text, extracting key information from documents, and performing sentiment analysis to learn public opinion. The release of OpenAI’s GPT-3 in 2020, with its 175 billion parameters, significantly boosted public interest due to its remarkable performance across these diverse applications~\cite{brown2020language}.

However, the knowledge in LLMs is frozen in their parameters at the time of training, leading to several limitations. These models tend to generate inaccurate or nonsensical information (hallucinations), need more detailed expertise in specific domains, particularly regarding new knowledge that emerges after their training and is often unclear in their decision-making processes (lack of interpretability).
Numerous research efforts have been dedicated to incorporating alternative knowledge sources, such as linguistic, retrieval-based, and graph-based knowledge, to enhance language models \cite{wei2021knowledge, zhen2022survey, hu2023survey}. 
These types of models are termed knowledge-enhanced pre-trained models (KEPLMs). In 2021, Wei et al. \cite{wei2021knowledge} surveyed various KEPLMs and their improved performance over vanilla pre-trained models (PLMs).
In 2022, Zhen et al. \cite{zhen2022survey} classified knowledge enhancement approaches into explicit and implicit incorporation methods.
While explicit methods inserted relevant knowledge into LLMs by modifying model inputs
and employing external memories, implicit methods focused on the knowledge contained within LLMs from training, like in BERT \cite{devlin2019bert}, which understood the contextual knowledge of words.
In the next year, Hu et al. \cite{hu2023survey} surveyed KEPLMs, focusing on two key tasks in NLP: Natural Language Understanding and Natural Language Generation.
While these earlier works touched on diverse external knowledge sources, a more recent review in 2024 by Yang et al. \cite{yang2024give} focused solely on injected knowledge from knowledge graphs (KGs). 
There has been a growing focus on KGs as sources of structured knowledge for LLM-based models. The intuitive structures of KGs effectively represent real-world knowledge by representing entities with nodes and relationships between them as edges, which enables a greater understanding of a word’s semantics via its context. 
Consequently, LLMs can effectively recall facts, particularly in specific domains. 

In fact, the utilization of Knowledge Graphs (KGs) is constrained by the availability of existing graphs.
KGs are difficult, costly, and time-consuming to construct, requiring numerous steps such as entity extraction, knowledge fusion, and coreference resolution. 
Moreover, KGs are specific to each domain, so separate ones are constructed for each application and may become irrelevant with time if not updated as knowledge evolves \cite{farghaly2024dkg}. 
Several works employ LLMs to enhance the KG construction process, but they use LLMs simply as information extractors. One example is a KG for the service domain BEAR \cite{yu2023bear}, which was created by prompting ChatGPT to extract content from unstructured data to populate an existing ontology. 

In previous surveys \cite{wei2021knowledge, zhen2022survey, hu2023survey, yang2024give}, there was an emphasis on utilising KGs as knowledge sources to support LLMs. 
KGs could provide external facts to LLMs, not necessarily only for LLMs' pre-training, but as retrieved facts to ground LLMs as well.
However, there was less focus on the benefits that LLMs could bring to KGs.
On the other hand, in methods like the previously mentioned BEAR \cite{yu2023bear}, LLMs were used merely to parse and extract relevant information from documents for KG construction, ignoring other benefits LLMs could bring to KG construction. 
The closest work to ours is a recent, comprehensive survey by Khorashadizadeh et al. \cite{khorashadizadeh2024research} outlining the mutual benefits between LLMs and KGs.
However, unlike previous surveys highlighting knowledge injection as the sole benefit KGs bring to LLMs, 
this paper will delve into the other benefits that KGs provide.
Furthermore, while \cite{khorashadizadeh2024research} classified LLM-KG cooperation methods by their uses, this paper seeks to explore the different ways that LLMs and KGs can be jointly used.

In fact, this integration
allows for better performance across a series of NLP tasks, such as named entity recognition and relation classification. 
Motivated by the diversity of ways KGs and LLMs can be employed in conjunction,
we propose the following research questions: 
\begin{itemize}
    \item (RQ1) How can KGs be utilised to enhance the capabilities of LLMs? 
    \item (RQ2) In what ways can LLMs be leveraged to support and enhance KGs
    \item (RQ3) Are there more advantages if
models combine KGs and LLMs in a more joint fashion?
\end{itemize}
To answer these questions,
we conducted an arXiv search for articles related to LLMs and KGs published no more than five years ago. 
We conducted our search from February 2024 to May 2024, and we chose arXiv as the main source of our review as it includes a wide range of articles. To identify relevant papers, we reviewed the title and abstract of each article by searching for keywords such as “Large Language Model” or “Knowledge Graph”, and we considered articles relevant if they reported on original research related to both LLMs and KGs, either with a topic on LLMs empowered by KGs, KGs empowered by LLMs or some Hybrid approaches.

To present the results of our search, we begin with a brief background of LLMs and KGs, including previous surveys, provided in Section \ref{background}.
The subsequent sections correspond to the research questions posed and detail methods including KGs added to LLMs in Section \ref{LLM-empowered-by-KG}, KGs empowered by LLMs in Section \ref{KG-empowered-by-LLM}, and some hybrid approaches in Section \ref{hybrid}. 
A thematic analysis is provided in Section \ref{thematic_analysis}, 
followed by discussions in Section \ref{strengths_and_lims} 
we review the described methodologies' different aspects, advantages, and limitations. 
Finally, the concluding remarks in Section \ref{conclusion} are where we close our review by presenting a general analysis and current and future challenges in this field. 

\section{Background}
\label{background}

\subsection{Large Language Models (LLMs)}

Various PLMs have been released in the past, including BERT \cite{devlin2019bert} and GPT-1 \cite{radford2018improving}. BERT (Bidirectional Encoder Representations from Transformers) was released in 2018, featuring a transformer-based model that understood contexts bidirectionally \cite{devlin2019bert}. The model could consider both the preceding and following words when processing input text, allowing it to accurately capture the meaning of words and sentences. The first GPT (Generative Pre-trained Transformer)  was also released in the same year, which focused on generating text by predicting the next word \cite{radford2018improving}.
Further developments saw the success of extending PLMs to LLMs by increasing their size and complexity, giving rise to their remarkable ability to comprehend natural language. 

LLMs 
nowadays are all based on the transformers architecture \cite{2311.07621}, which excels in handling long sequences due to its characteristic self-attention mechanism.
LLMs generally take a sequence of text or code called a prompt as the input. The tokenizer, a part of the LLM architecture, subsequently converts the input into a list of tokens and feeds them into the model, where each token is a word in the input prompt. 
New tokens are generated one by one until they reach a special token indicating the end of the generation, or the total length exceeds the limit. 
These generated tokens are converted back into the text format as the final output of the model. Figure \ref{fig: LLM text generation workflow} shows a simple scheme of this workflow. This process can be formally modelled through some equations. The input prompt, after being tokenized, can be seen as a list of tokens $x$, where $x$ = [$x_1$, . . . , $x_n$] if there are $n$ tokens in total. 
LLMs based on the transformer architecture usually maintain a sequence of hidden states. At step $t$, hidden state $h_t$ can be calculated using the current token $x_t$ and all the previous hidden states:

\begin{equation}
    h_t = LLM(x_t, [h_0,\ldots,h_{n-1}])
\end{equation}

The model then further transforms $h_t$ into a probability distribution, which can be used to sample the next generated token: 

\begin{equation}
    P(x) = \prod_{t=1}^{|x|} P(x_t|[h_0,\ldots,h_{t-1}])
\end{equation}

In this case, due to its ability to recognise, summarise, translate, predict, and generate text and other forms of content, current LLMs have a wide range of applications, including question answering and code generation.
More recently, some focus has been directed towards multimodal LLMs, which give LLMs abilities like the capabilities of vision in Google’s Gemini \cite{geminiteam2024gemini} and GPT-4 with vision (GPT-4V) \cite{openai2023gpt}.

\begin{center}
    \centering
    \includegraphics[height=1.5cm]{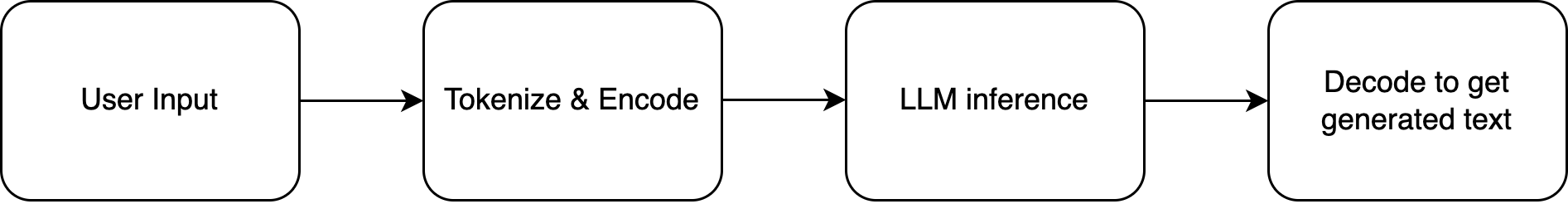}
    \captionof{figure}{LLM text generation workflow.}
    \label{fig: LLM text generation workflow}
\end{center}

\subsection{Knowledge Graphs (KGs)}

A KG is a directed labelled graph in which nodes represent any real-world entity or concept, and edges represent the relationships between nodes. This structured data format has proven effective and applicable to a wide variety of domains, including biology and finance, as well as in modelling social networks or storing general information like in the Google Knowledge Graph. Aside from being able to model relationships, KGs can provide further insight into a word’s semantics via its context or neighbouring nodes. The graph structure can also be studied, as was done in \cite{aparicio2024emerging} to uncover research trends over time, or in \cite{wang2024acemap} where structure provided insight into the impact and success of research publications. Data in KGs is usually presented as a (subject, object, predicate) triple. This could be extended in the case of temporal knowledge graphs, which use a quadruple representation (subject, object, predicate, timestamp) to capture facts over time.

The construction of a KG involves three general steps: knowledge acquisition, knowledge refinement, and knowledge evolution \cite{zhong2023comprehensive}. 
Knowledge acquisition involves collecting information about entities and relations from multi-structured data to build the KG. 
Since extracted triples could be incomplete, the next knowledge refinement step fixes these issues with additional data. 
Finally, the evolution of real-world knowledge over time may not be reflected in the static KGs that were built, so the graphs are dynamically updated with the knowledge evolution step.
KG construction methods are based on crowdsourcing or text mining. Crowdsourcing-based KGs like WordNet \cite{fellbaum1998wordnet} and ConceptNet \cite{speer2017conceptnet} require significant human labour to construct as they depend on contributions from volunteers. Meanwhile, KGs constructed by text mining utilise a series of subtasks like named entity recognition and relationship extraction to extract graph data from text. 
The resulting KGs are, however, limited by the quality and scope of the given data.
Both crowdsourcing and text mining methods of construction suffer from limitations.
As such, numerous methods have been proposed to utilise LLMs for tasks like relation extraction and property identification in the KG construction process, as discussed in \cite{khorashadizadeh2024research}.
Utilising LLMs, KG construction could be more automatic whilst maintaining accuracy.




\section{LLMs Empowered by KGs} \label{LLM-empowered-by-KG}
\textbf{Knowledge injection.}
A myriad of techniques have been implemented in research and industry to perform knowledge injection using KGs, usually including additional knowledge in LLM prompts as shown in Figure \ref{fig: KG-enhanced LLMs.}. For instance, Baek et al. \cite{ baek2023knowledgeaugmented} proposed KAPING (Knowledge-Augmented language model PromptING), which retrieved facts from a KG and prepended them to input questions to construct LLM prompts for zero-shot question answering. A similar approach was adopted in Sen et al. \cite{sen2023knowledge} where, instead of the previous approach, the facts from the KG were weighted by a Knowledge Graph Question Answering (KGQA) before being fed into the LLM. In KICGPT (Knowledge In Context with GPT) \cite{wei2023kicgpt}, retrieved KG facts were re-ranked by the LLM. An application example is DRAK (Domain-specific Retrieval-Augmented Knowledge), where retrieved KG facts are also useful to LLMs in the biomolecular domain, which requires structured knowledge \cite{liu2024drak}. These approaches are a mere subset of the numerous variations of techniques utilised to inject KG knowledge into LLM prompts.
Rather than including retrieved KG facts into LLM prompts, Knowledge Solver \cite{feng2023knowledge} teaches LLMs to traverse KGs in a multi-hop way to reason the answer to a question. In this way, KGs could provide facts that LLMs could reason over, grounding them in the process.
\textbf{Increasing LLM explainability.}
On the other hand, KGs can contribute more to LLMs than by simply providing facts for knowledge grounding. For the question-answering task, QA-GNN (Question Answering Graph Neural Network) \cite{yasunaga2022qagnn} performed joint reasoning over a LLM encoding of the question context and KG to unify the two representations. For better model interpretability, a graph neural network (GNN) was used to calculate weights between graph nodes, providing a path of reasoning that the model took through the KG to get to the answer. Another example is LMExplainer \cite{chen2023lmexplainer}, which used a KG and graph attention neural network to understand key decision signals of LLMs, which were converted into natural language explanations for better explainability. KGs could, therefore, also allow for better interpretability of LLMs and offer insights into LLMs’ reasoning processes, which in turn increase humans’ trust in LLMs.

\textbf{Semantic understanding.}
KGs can also be applied to add semantic understanding or entity embeddings into LLMs. For instance, LUKE (Language Understanding with Knowledge-based Embeddings) \cite{yamada2020luke}, as an extension of BERT, is an entity-aware self-attention mechanism that can help the model treat words and entities in a given text as independent tokens and output contextualised representations of them. As for adding semantic understanding, a recent methodology called Right for Right Reasons (R3) \cite{toroghi2024right} for performing KGQA using LLMs casts the problem of common sense KGQA as a tree-structured search to make full use of surfaced commonsense axioms, a key property that makes the reasoning procedure verifiable, such that semantic understandings from KGs can be added into LLMs.

\begin{center}
    \centering
    \includegraphics[height=4.5cm]{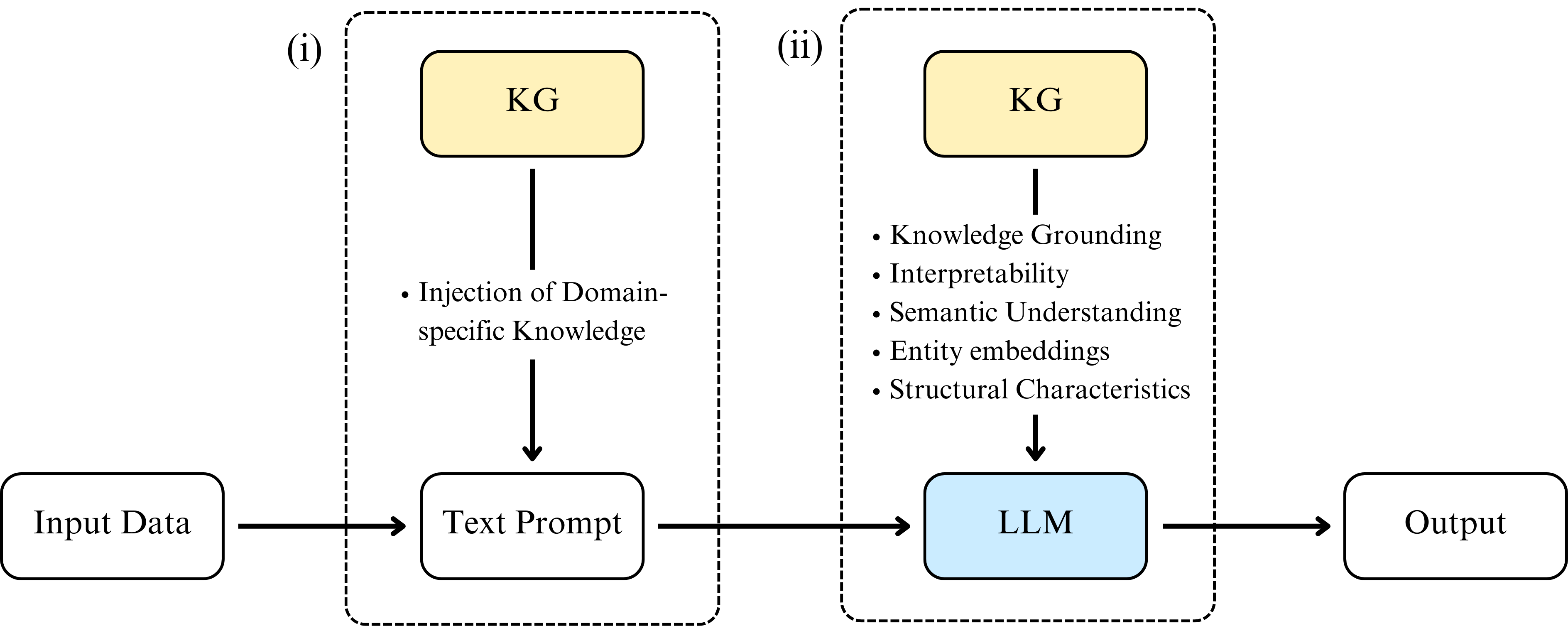}
    \captionof{figure}{KG-enhancement for LLMs can come in the form of (i) KG knowledge injection into LLM prompts or (ii) other methods where KGs directly contribute to LLMs.}
    \label{fig: KG-enhanced LLMs.}
\end{center}

\section{KGs Empowered by LLMs} \label{KG-empowered-by-LLM}
There are numerous examples of LLMs being empowered by KGs, but one could also consider the opposite direction for integration: KGs empowered by LLMs. 

\textbf{Temporal forecasting.}
Recent research has shown that LLMs can perform forecasting with KG data, especially for a special subset of KGs having directions and timestamps, namely Temporal Knowledge Graphs (TKGs). Most advanced research on TKGs mainly focuses on predicting future facts when given historical facts, where using LLMs can be particularly beneficial. For instance, Xia and their colleagues propose a Chain-of-History (CoH) reasoning method for TKG prediction \cite{xia2024enhancing}, where an LLM is mainly used to understand the semantic meaning of entities, relationships, and timestamps in a TKG by exploring important high-order history chains step-by-step and reasoning the answers to the query only based on inferred history chains in the last step. Alternatively, by using in-context learning (ICL) with LLMs whereby a few examples were provided to the LLM so it could learn to perform forecasting, Lee et al. \cite{lee2023temporal} fed TKG facts to the LLM and found that LLMs were surprisingly capable of learning patterns from historical data. 
This is despite the lack of special architectures or modules usually required to perform this KG task. The ability of LLMs to perform what is typically a KG task allows for the possibility of using natural language to perform forecasting.

\textbf{Knowledge graph construction.}
As discussed previously, one major challenge is the time-consuming and costly construction process of KGs, another aspect to which LLMs can contribute in various ways as depicted in Figure \ref{fig: LLM-enhanced KG construction.}. 
LLMs are trained on large, diverse datasets and store this knowledge implicitly. BertNet \cite{hao2023bertnet} sought to harvest KGs of arbitrary relations from LLMs, useful for general KGs. To achieve this, an initial prompt was paraphrased several times and the LLM would provide responses to each of the paraphrased prompts, which were converted into entity pairs and ranked. The top-ranking pairs formed the KG. Kommineni et al. \cite{kommineni2024human} crafted a semi-automatic KG construction pipeline utilising ChatGPT-3.5, which prompted the LLM to generate high-level competency questions about the data. The LLM was instructed to extract entities and relationships from these questions to form an ontology, then map retrieved information from documents onto the ontology to construct the KG. Similar examples include AutoRD \cite{cao2024autord}, a useful framework introduced recently for extracting information about rare diseases and constructing corresponding knowledge graphs. This system can process unstructured medical text as input and output extraction results and a knowledge graph, where LLM is used to extract entities and relations from medical ontologies. Most recently, an unsupervised framework called TKGCon (Theme-specific Knowledge Graph Construction) \cite{ding2024automated} utilised LLMs to construct both ontologies and theme-specific KGs, by relying on LLMs to generate and decide relations between entities to construct graph edges. These methods signal that LLMs are capable of more than knowledge extraction from unstructured data. They can also process and reason over data to construct and complete KGs.
Furthermore, Khorashadizadeh et al. \cite{khorashadizadeh2024research} outlined other methods that used LLMs for specific KG construction tasks like text-to-ontology mapping, entity extraction, and ontology alignment.
LLMs were also used for KG validation through fact-checking and inconsistency detection.

\begin{center}
    \centering
    \includegraphics[height=5cm]{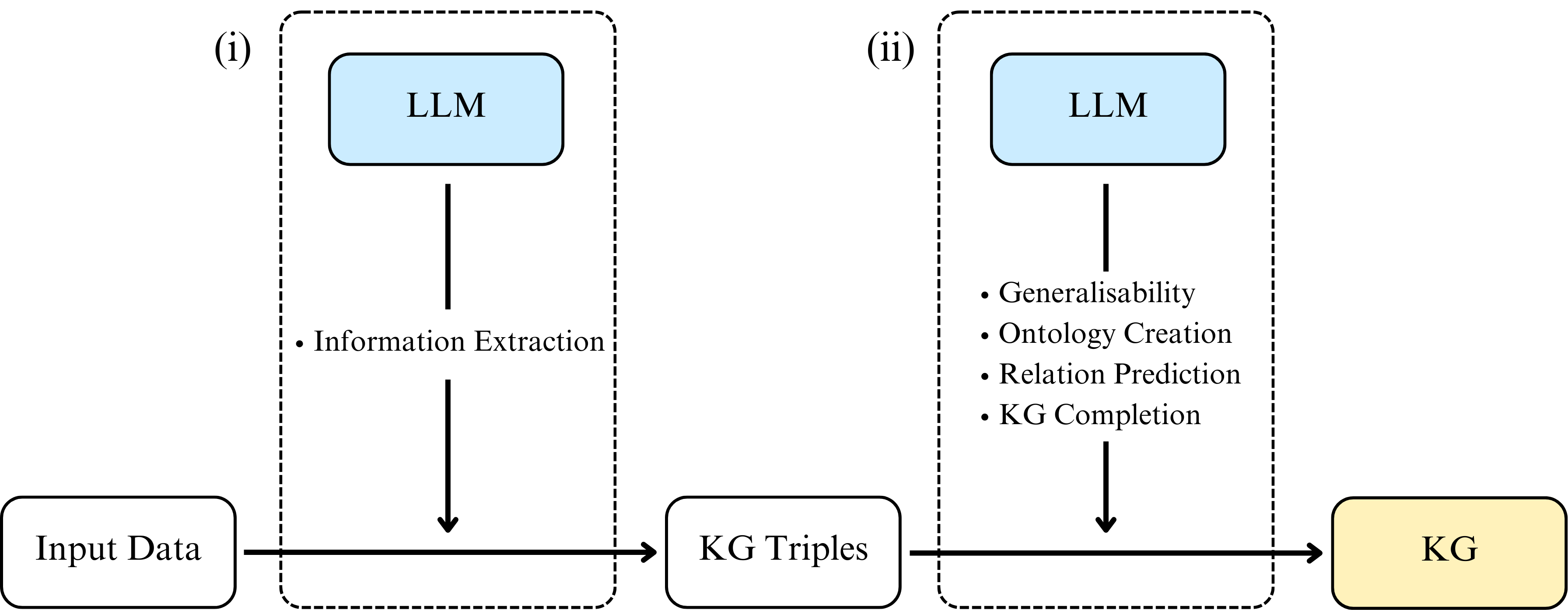}
    \captionof{figure}{LLM-enhanced KG construction, where (i) LLMs are used for information extraction or (ii) for general KG data operations.}
    \label{fig: LLM-enhanced KG construction.}
\end{center}

\section{Hybrid Approaches} \label{hybrid}

\textbf{Fusing textual and knowledge embeddings.}
In contrast to the methods presented in previous sections, the approaches explained in this section combine KGs and LLMs in a more unified way to build upon the explicit knowledge from KGs and implicit knowledge found within LLMs. 
An example of this is ERNIE (Enhanced Language RepresentatioN with Informative Entities) \cite{zhang2019ernie}, which fuses lexical, syntactic, and knowledge information together by stacking a textual T-Encoder with a knowledge K-Encoder to represent word tokens and entities in a unified feature space,
similar to the illustration in Figure \ref{fig: Hybrid Approaches.}. 
The T-Encoder functions exactly as the LLM, BERT \cite{devlin2019bert}, to obtain a feature representation of word tokens.
Together with entity embeddings from KGs, these features are subsequently fed into the K-Encoder, which fuses these separate embeddings into a unified output embedding.
This approach improved performance for knowledge-driven tasks like entity typing and relation classification. 
A similar architecture is adopted by CokeBERT \cite{su2021cokebert}, which utilises an LLM to encode word tokens and extracts knowledge contexts from KGs for each entity detected in text. The word and knowledge embeddings were fused using the K-Encoder from \cite{zhang2019ernie}.

\textbf{Adding knowledge within the LLM.}
Other approaches fuse KG knowledge with textual embeddings within the LLM. KnowBERT \cite{peters2019knowledge} includes an additional KAR (Knowledge Attention and Recontextualization) component to BERT’s architecture. It takes contextual representations from a BERT layer, computes a knowledge-enhanced representation using a list of possible entity links from a KG, and passes the new representation to the next transformer block in BERT. Compared to standard BERT, KnowBERT exhibited better performance on relation extraction, words in context, and entity typing tasks. 

\textbf{Multimodal LLMs.}
Both types of embeddings have also been combined to perform visual question-answering. KRISP \cite{marino2021krisp} uses a multimodal BERT-pretrained transformer to process a question and image pair in its implicit knowledge model. A separate explicit knowledge model constructs a KG from question and image symbols, including places, objects, and attributes. The two models work together to predict answers regarding the image.

A common theme amongst these approaches is that they perform better tasks like entity typing, which involves assigning entities into categories based on their semantic meanings, and visual question answering. These tasks require models to better understand semantics, which appears to be an advantage of the hybrid approaches.

\begin{center}
    \centering
    \includegraphics[height=7.5cm]{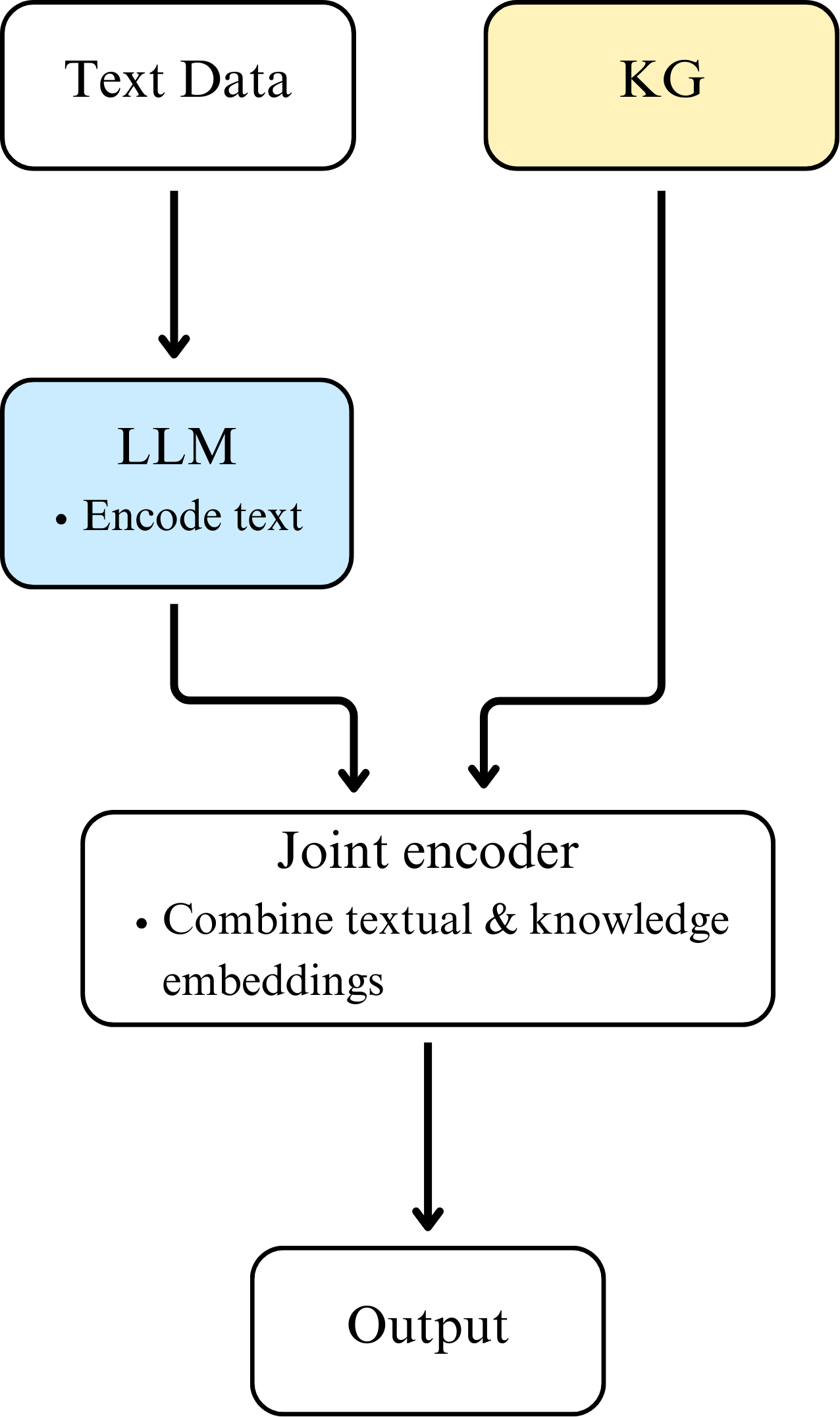}
    \captionof{figure}{Hybrid Approaches that combine text and KG embeddings.}
    \label{fig: Hybrid Approaches.}
\end{center}

\section{Thematic Analysis} \label{thematic_analysis}

In this section, the important models mentioned so far are categorised into "Add-ons" versus "Joint" Knowledge Graph + LLM approaches. 
\begin{itemize}
    \item Add-ons - 
    The models categorised here use LLMs and KGs as supplementary tools to enhance their functionality. They include models where LLMs are used for creating KGs or KGs are used for providing information to LLMs. The purpose behind employing this approach is so that KGs and LLMs can operate independently to maximizing qualities such as scalability, cost reduction, or flexibility. 
    \item Joint - 
    The models under this category leverage the combined strengths of LLMs and KGs to achieve enhanced performance in specific tasks. The tasks are application-dependent, and this approach can provide comprehensive understanding, optimized results, and improved accuracy.
\end{itemize}

The rationale behind categorizing each model into a specific category, along with the potential advantages of the particular approach for the model, has been provided in the Explanation section of Table \ref{table:table1}.
This is purely our contribution. This novel perspective can be used to understand how these technologies can be used in isolation or together. This is an angle that has not been previously explored in other papers. 

\setcounter{table}{0}

\begin{longtable}
{|p{2cm}|p{3cm}|p{10.3cm}|}
  \hline
  \vspace{0pt}Model & \vspace{0pt}Add-on or Joint & \vspace{0pt}Explanation \\
  \hline
  \vspace{1pt}KnowPhish & \vspace{1pt}Add-on & \vspace{1pt} 
  KnowPhish Detector (KPD) combines:
  \begin{itemize}
      \setlength\itemsep{0pt} 
      \setlength\parskip{0pt} 
      \item KnowPhish: For logo-based detection. This on its own is just a Brand Knowledge Base.
      \item Large Language Model (LLM): To extract brand information from webpage text, enabling detection of phishing attempts even without logos.
  \end{itemize}
  \vspace{2pt} 
    Wikidata, a knowledge graph, was utilised to create the Brand Knowledge Base by providing information about brands as entities, including details such as logos, official website URLs, and aliases.
  The KG and LLM are used for two separate procedures and two separate intentions and act as add-ons
  \cite{li2024knowphish}. The KG allows for better scaling to extend to many brands, while the LLM allows for extracting brand information from webpages from text. Using the KGs and LLMs as add-ons offers improved detection accuracy. 
  \\
  \hline
  \vspace{1pt}BEAR & \vspace{1pt}Add-on & \vspace{1pt} 
  BEAR is an innovative open Knowledge Graph designed to capture knowledge pertinent to the service computing community. Service computing aims to act as a bridge between business services and IT services 
  \cite{yu2023bear}.
An ontology specific to the domain serves as the foundation for the KG, outlining the concepts and characteristics that would later populate the graph. The LLM is only used to improve the data extraction process to update the KG with the required relevant information.
Utilising the LLM as an add-on eliminates the need for manual data annotation, saving time and costs.
 \\
  \hline
   \vspace{1pt}K-BERT & \vspace{1pt}Joint  & \vspace{1pt} 
  BERT is an LLM that lacks domain-specific knowledge since it is pre-trained in general language from 
large-scale corpora. K-BERT addresses this by injecting domain knowledge from knowledge graphs into sentences.  By using the functionalities of LLMs and KGs jointly,  good performance in domain-specific tasks can be achieved in K-BERT without requiring extensive pre-training
\cite{liu2019kbert}.
 \\
  \hline
    \vspace{1pt}ERNIE & \vspace{1pt}Joint  & \vspace{1pt} 
 ERNIE is a language representation model trained on large-scale textual corpora and KGs, allowing it to simultaneously utilise lexical, syntactic, and knowledge information. Incorporating KGs in a joint fashion results in better language understanding
\cite{zhang2019ernie}.  
 \\
  \hline
    \vspace{1pt}LMExplainer & \vspace{1pt}Joint  & \vspace{1pt} 
 LMExplainer is a knowledge-enhanced tool designed to explain the predictions made by LLMs. It uses KGs and graph attention neural networks to describe the reasoning behind the model's predictions
\cite{chen2023lmexplainer}. This joint approach ensures that the explanations are human-understandable.
 \\
  \hline
 \caption{Models and types of interaction among KG-LLMs.}
\label{table:table1} 
\end{longtable}

\section{Strengths and Limitations of Existing Research} \label{strengths_and_lims}

This section reviews the strengths and limitations of existing research covered in this paper, which can be crucial to understanding this joint approach of combining KGs and LLMs. One of the significant strengths we have identified is the performance improvement brought by the utilisation of KGs and LLMs in a joint fashion, especially in the knowledge-driven domain. Models combining KGs and LLMs typically display a better semantic understanding of knowledge, thus enabling them to perform tasks like entity typing better.
Additionally, as was seen in several methods like QA-GNN \cite{yasunaga2022qagnn}, interpretability and explainability of the model can be increased when combining KGs with LLMs, which are particularly important factors when LLMs are being adopted in sensitive domains like healthcare, education, and emergency response.
\\

Nevertheless, current research also has limitations that may hinder this joint approach's broader application or effectiveness. One of the major issues is that KGs in some domains may not be widely available, thus limiting the ability to integrate KGs and LLMs. Even if LLMs were employed to help automate the KG construction process, they could hallucinate or produce incorrect results, compromising KG data's accuracy and validity.
\\

Furthermore, integrating KGs and LLMs can lead to even larger parameter sizes and longer running times. Moreover, if additional modules are included, as was the case in several hybrid models discussed in Section \ref{hybrid}, the extra time and computational resources would be needed to train these modules as well. As demonstrated by Yang et al. \cite{yang2024facts}, due to these knowledge graph-enhanced pre-trained language models (KGPLMs) injecting the knowledge encoder module into pre-trained language models (PLMs), their running times are consistently longer than a vanilla LLM, BERT, across pre-training, fine-tuning, and inference stages. 
This is although incorporating external knowledge from KGs makes it easier for them to train and enhances their performance.
\\


Another challenge is that KGs and LLMs will likely suffer from becoming outdated due to the rapid evolution of knowledge. One may need to update KGs or LLMs frequently to mitigate this issue. KGs are significantly easier than LLMs to update, although additional KG completion steps are required. In the case of LLMs, the impracticality of repeating lengthy and costly training processes can significantly affect the time and costs involved.
This calls for additional methods to update LLMs, whether it is via KGs or other knowledge sources.
\\

The effectiveness of KG and LLM integration is also an area that demands further research. According to a study from \cite{hou2022enhanced}, only a marginal amount of knowledge is successfully integrated into two well-known knowledge-enhanced Language Models (LM), namely ERNIE and K-Adapter, and simply increasing the size of the Knowledge Integration corpus may not lead to better knowledge-enhanced LMs. This highlights a critical gap in the current approach and suggests a need for more effective integration methods.


\section{Conclusion} \label{conclusion}
Driven by research questions “How can KGs be utilised to enhance the capabilities of LLMs?” (RQ1), "In what ways can LLMs be leveraged to support and enhance KGs" (RQ2) and “Are there more advantages if models combine KGs and LLMs in a more joint fashion” (RQ3), we conducted a quick review to explore the prevalence of using KGs purely for knowledge injection to support LLM-based models, and of using LLMs merely as information extractors to support KG-based models. To understand the advantages of this joint approach, we reviewed over 20 state-of-the-art articles from arXiv and classified existing methods into three different categories, including KGs empowered by LLMs (KGs adding interpretability, semantic understanding, and entity embeddings into LLMs), LLMs empowered by KGs (LLMs do forecasting with KG data, injecting implicit knowledge, and contributing to KG construction) and some Hybrid Approaches where KGs and LLMs are combined in a more unified way. 
We also provided a Thematic Analysis for these methods and discussed their strengths and limitations. To answer research questions, we found that models typically use KGs or LLMs like add-ons, and there are more advantages if models combine KGs and LLMs in a more joint fashion. Additionally, we found that even though the joint approach can provide significant improvement in the performance of the model by increasing its interpretability or explainability, current research also has its limitations, such as limited domains for KGs, higher consumption of computational resources, outdated frequently due to the fast evolution of knowledge, and the lack of effectiveness in Knowledge Integration. \\

This research area of combining KGs with LLMs represents a critical part in the rising trend of artificial intelligence (AI), and can potentially lead to more reliable and context-aware AI systems. Such models are equipped with domain-specific knowledge and contribute to a wider range of applications than solely using KGs or LLMs for problem-solving. Eventually, this research area will significantly impact how we construct a more robust and explainable AI system with higher performance and can provide avenues for other future development.

Although the joint approach of combining KGs with LLMs has succeeded, some unsolved challenges remain. To address the low effectiveness of knowledge integration, future studies may continue exploring the potential solution to this problem by modifying model architecture or fine-tuning. One possible solution could be to inject knowledge into feature-based pre-training models.
Future studies could also focus on developing a smaller integrated model to reduce computational resources and time, as integrating KGs and LLMs typically leads to larger parameter sizes and longer running time. Given the fact that a smaller KGPLM can outperform a larger plain LLM, it is possible that an optimal integrated model that requires fewer computational resources can be achieved.
In the past year, there has also been a surge in interest in multimodal LLMs, which can process audio, image, or video data together with text, with a handful of multimodal LLMs released each month since the start of 2023. As these models are built on LLM backbones, it is foreseeable that they could also inherit some of the limitations LLMs have experienced thus far and might benefit from incorporating KGs. As such, other studies could explore the potential use of multimodal KGs when combined with LLMs for the recent advances in the research of multimodal models.

\bibliographystyle{unsrturl}  
\bibliography{references}  

\begin{thebibliography}{10}

\bibitem{devlin2019bert}
Jacob Devlin, Ming-Wei Chang, Kenton Lee, and Kristina Toutanova.
\newblock Bert: Pre-training of deep bidirectional transformers for language understanding, 2019.
\newblock \href {https://arxiv.org/abs/1810.04805} {\path{arXiv:1810.04805}}.

\bibitem{raffel2023exploringlimitstransferlearning}
Colin Raffel, Noam Shazeer, Adam Roberts, Katherine Lee, Sharan Narang, Michael Matena, Yanqi Zhou, Wei Li, and Peter~J. Liu.
\newblock Exploring the limits of transfer learning with a unified text-to-text transformer, 2023.
\newblock URL: \url{https://arxiv.org/abs/1910.10683}, \href {https://arxiv.org/abs/1910.10683} {\path{arXiv:1910.10683}}.

\bibitem{radford2018improving}
Alec Radford, Karthik Narasimhan, Tim Salimans, Ilya Sutskever, et~al.
\newblock Improving language understanding by generative pre-training.
\newblock 2018.
\newblock URL: \url{https://api.semanticscholar.org/CorpusID:49313245}.

\bibitem{brown2020language}
Tom Brown, Benjamin Mann, Nick Ryder, Melanie Subbiah, Jared~D Kaplan, Prafulla Dhariwal, Arvind Neelakantan, Pranav Shyam, Girish Sastry, Amanda Askell, et~al.
\newblock Language models are few-shot learners.
\newblock {\em Advances in neural information processing systems}, 33:1877--1901, 2020.

\bibitem{wei2021knowledge}
Xiaokai Wei, Shen Wang, Dejiao Zhang, Parminder Bhatia, and Andrew Arnold.
\newblock Knowledge enhanced pretrained language models: A compreshensive survey, 2021.
\newblock \href {https://arxiv.org/abs/2110.08455} {\path{arXiv:2110.08455}}.

\bibitem{zhen2022survey}
Chaoqi Zhen, Yanlei Shang, Xiangyu Liu, Yifei Li, Yong Chen, and Dell Zhang.
\newblock A survey on knowledge-enhanced pre-trained language models, 2022.
\newblock \href {https://arxiv.org/abs/2212.13428} {\path{arXiv:2212.13428}}.

\bibitem{hu2023survey}
Linmei Hu, Zeyi Liu, Ziwang Zhao, Lei Hou, Liqiang Nie, and Juanzi Li.
\newblock A survey of knowledge enhanced pre-trained language models.
\newblock {\em IEEE Transactions on Knowledge and Data Engineering}, 36(4):1413--1430, 2024.
\newblock \href {https://doi.org/10.1109/TKDE.2023.3310002} {\path{doi:10.1109/TKDE.2023.3310002}}.

\bibitem{yang2024give}
Linyao Yang, Hongyang Chen, Zhao Li, Xiao Ding, and Xindong Wu.
\newblock Give us the facts: Enhancing large language models with knowledge graphs for fact-aware language modeling.
\newblock {\em IEEE Transactions on Knowledge and Data Engineering}, pages 1--20, 2024.
\newblock \href {https://doi.org/10.1109/TKDE.2024.3360454} {\path{doi:10.1109/TKDE.2024.3360454}}.

\bibitem{farghaly2024dkg}
Maha Farghaly, Mahmoud Mounir, Mostafa Aref, and Sherin~M. Moussa.
\newblock Investigating the challenges and prospects of construction models for dynamic knowledge graphs.
\newblock {\em IEEE Access}, 12:40973--40988, 2024.
\newblock \href {https://doi.org/10.1109/ACCESS.2024.3378514} {\path{doi:10.1109/ACCESS.2024.3378514}}.

\bibitem{yu2023bear}
Shuang Yu, Tao Huang, Mingyi Liu, and Zhongjie Wang.
\newblock Bear: Revolutionizing service domain knowledge graph construction with llm.
\newblock In Flavia Monti, Stefanie Rinderle-Ma, Antonio Ruiz~Cort{\'e}s, Zibin Zheng, and Massimo Mecella, editors, {\em Service-Oriented Computing}, pages 339--346, Cham, 2023. Springer Nature Switzerland.

\bibitem{khorashadizadeh2024research}
Hanieh Khorashadizadeh, Fatima~Zahra Amara, Morteza Ezzabady, Frédéric Ieng, Sanju Tiwari, Nandana Mihindukulasooriya, Jinghua Groppe, Soror Sahri, Farah Benamara, and Sven Groppe.
\newblock Research trends for the interplay between large language models and knowledge graphs, 2024.
\newblock \href {https://arxiv.org/abs/2406.08223} {\path{arXiv:2406.08223}}.

\bibitem{2311.07621}
Micaela~E. Consens, Cameron Dufault, Michael Wainberg, Duncan Forster, Mehran Karimzadeh, Hani Goodarzi, Fabian~J. Theis, Alan Moses, and Bo~Wang.
\newblock To transformers and beyond: Large language models for the genome, nov 2023.
\newblock URL: \url{http://arxiv.org/abs/2311.07621v1}, \href {https://arxiv.org/abs/2311.07621} {\path{arXiv:2311.07621}}.

\bibitem{geminiteam2024gemini}
Gemini Team.
\newblock Gemini: A family of highly capable multimodal models, 2024.
\newblock \href {https://arxiv.org/abs/2312.11805} {\path{arXiv:2312.11805}}.

\bibitem{openai2023gpt}
R~OpenAI.
\newblock Gpt-4v (ision) system card.
\newblock {\em Citekey: gptvision}, 2023.

\bibitem{aparicio2024emerging}
Joao~T. Aparicio, Elisabete Arsenio, Francisco Santos, and Rui Henriques.
\newblock Using dynamic knowledge graphs to detect emerging communities of knowledge.
\newblock {\em Knowledge-Based Systems}, 294:111671, 2024.
\newblock URL: \url{https://www.sciencedirect.com/science/article/pii/S095070512400306X}, \href {https://doi.org/10.1016/j.knosys.2024.111671} {\path{doi:10.1016/j.knosys.2024.111671}}.

\bibitem{wang2024acemap}
Xinbing Wang, Luoyi Fu, Xiaoying Gan, Ying Wen, Guanjie Zheng, Jiaxin Ding, Liyao Xiang, Nanyang Ye, Meng Jin, Shiyu Liang, Bin Lu, Haiwen Wang, Yi~Xu, Cheng Deng, Shao Zhang, Huquan Kang, Xingli Wang, Qi~Li, Zhixin Guo, Jiexing Qi, Pan Liu, Yuyang Ren, Lyuwen Wu, Jungang Yang, Jianping Zhou, and Chenghu Zhou.
\newblock Acemap: Knowledge discovery through academic graph, 2024.
\newblock \href {https://arxiv.org/abs/2403.02576} {\path{arXiv:2403.02576}}.

\bibitem{zhong2023comprehensive}
Lingfeng Zhong, Jia Wu, Qian Li, Hao Peng, and Xindong Wu.
\newblock A comprehensive survey on automatic knowledge graph construction, 2023.
\newblock \href {https://arxiv.org/abs/2302.05019} {\path{arXiv:2302.05019}}.

\bibitem{fellbaum1998wordnet}
Christiane Fellbaum.
\newblock {\em WordNet: An electronic lexical database}.
\newblock MIT press, 1998.

\bibitem{speer2017conceptnet}
Robyn Speer, Joshua Chin, and Catherine Havasi.
\newblock Conceptnet 5.5: An open multilingual graph of general knowledge.
\newblock In {\em Proceedings of the AAAI Conference on Artificial Intelligence}, volume~31, 2017.
\newblock URL: \url{http://dx.doi.org/10.1609/aaai.v31i1.11164}, \href {https://doi.org/10.1609/aaai.v31i1.11164} {\path{doi:10.1609/aaai.v31i1.11164}}.

\bibitem{baek2023knowledgeaugmented}
Jinheon Baek, Alham~Fikri Aji, and Amir Saffari.
\newblock Knowledge-augmented language model prompting for zero-shot knowledge graph question answering, 2023.
\newblock \href {https://arxiv.org/abs/2306.04136} {\path{arXiv:2306.04136}}.

\bibitem{sen2023knowledge}
Priyanka Sen, Sandeep Mavadia, and Amir Saffari.
\newblock Knowledge graph-augmented language models for complex question answering.
\newblock In {\em Proceedings of the 1st Workshop on Natural Language Reasoning and Structured Explanations (NLRSE)}, pages 1--8, 2023.

\bibitem{wei2023kicgpt}
Yanbin Wei, Qiushi Huang, Yu~Zhang, and James Kwok.
\newblock Kicgpt: Large language model with knowledge in context for knowledge graph completion.
\newblock In {\em Findings of the Association for Computational Linguistics: EMNLP 2023}. Association for Computational Linguistics, 2023.
\newblock URL: \url{http://dx.doi.org/10.18653/v1/2023.findings-emnlp.580}, \href {https://doi.org/10.18653/v1/2023.findings-emnlp.580} {\path{doi:10.18653/v1/2023.findings-emnlp.580}}.

\bibitem{liu2024drak}
Jinzhe Liu, Xiangsheng Huang, Zhuo Chen, and Yin Fang.
\newblock Drak: Unlocking molecular insights with domain-specific retrieval-augmented knowledge in llms.
\newblock {\em Authorea Preprints}, 2024.

\bibitem{feng2023knowledge}
Chao Feng, Xinyu Zhang, and Zichu Fei.
\newblock Knowledge solver: Teaching llms to search for domain knowledge from knowledge graphs, 2023.
\newblock \href {https://arxiv.org/abs/2309.03118} {\path{arXiv:2309.03118}}.

\bibitem{yasunaga2022qagnn}
Michihiro Yasunaga, Hongyu Ren, Antoine Bosselut, Percy Liang, and Jure Leskovec.
\newblock Qa-gnn: Reasoning with language models and knowledge graphs for question answering, 2022.
\newblock \href {https://arxiv.org/abs/2104.06378} {\path{arXiv:2104.06378}}.

\bibitem{chen2023lmexplainer}
Zichen Chen, Ambuj~K Singh, and Misha Sra.
\newblock Lmexplainer: a knowledge-enhanced explainer for language models, 2023.
\newblock \href {https://arxiv.org/abs/2303.16537} {\path{arXiv:2303.16537}}.

\bibitem{yamada2020luke}
Ikuya Yamada, Akari Asai, Hiroyuki Shindo, Hideaki Takeda, and Yuji Matsumoto.
\newblock Luke: Deep contextualized entity representations with entity-aware self-attention, 2020.
\newblock \href {https://arxiv.org/abs/2010.01057} {\path{arXiv:2010.01057}}.

\bibitem{toroghi2024right}
Armin Toroghi, Willis Guo, Mohammad Mahdi~Abdollah Pour, and Scott Sanner.
\newblock Right for right reasons: Large language models for verifiable commonsense knowledge graph question answering, 2024.
\newblock \href {https://arxiv.org/abs/2403.01390} {\path{arXiv:2403.01390}}.

\bibitem{xia2024enhancing}
Yuwei Xia, Ding Wang, Qiang Liu, Liang Wang, Shu Wu, and Xiaoyu Zhang.
\newblock Enhancing temporal knowledge graph forecasting with large language models via chain-of-history reasoning, 2024.
\newblock \href {https://arxiv.org/abs/2402.14382} {\path{arXiv:2402.14382}}.

\bibitem{lee2023temporal}
Dong-Ho Lee, Kian Ahrabian, Woojeong Jin, Fred Morstatter, and Jay Pujara.
\newblock Temporal knowledge graph forecasting without knowledge using in-context learning, 2023.
\newblock \href {https://arxiv.org/abs/2305.10613} {\path{arXiv:2305.10613}}.

\bibitem{hao2023bertnet}
Shibo Hao, Bowen Tan, Kaiwen Tang, Bin Ni, Xiyan Shao, Hengzhe Zhang, Eric~P. Xing, and Zhiting Hu.
\newblock Bertnet: Harvesting knowledge graphs with arbitrary relations from pretrained language models, 2023.
\newblock \href {https://arxiv.org/abs/2206.14268} {\path{arXiv:2206.14268}}.

\bibitem{kommineni2024human}
Vamsi~Krishna Kommineni, Birgitta König-Ries, and Sheeba Samuel.
\newblock From human experts to machines: An llm supported approach to ontology and knowledge graph construction, 2024.
\newblock \href {https://arxiv.org/abs/2403.08345} {\path{arXiv:2403.08345}}.

\bibitem{cao2024autord}
Lang Cao, Jimeng Sun, and Adam Cross.
\newblock Autord: An automatic and end-to-end system for rare disease knowledge graph construction based on ontologies-enhanced large language models, 2024.
\newblock \href {https://arxiv.org/abs/2403.00953} {\path{arXiv:2403.00953}}.

\bibitem{ding2024automated}
Linyi Ding, Sizhe Zhou, Jinfeng Xiao, and Jiawei Han.
\newblock Automated construction of theme-specific knowledge graphs.
\newblock {\em arXiv preprint arXiv:2404.19146}, 2024.

\bibitem{zhang2019ernie}
Zhengyan Zhang, Xu~Han, Zhiyuan Liu, Xin Jiang, Maosong Sun, and Qun Liu.
\newblock Ernie: Enhanced language representation with informative entities, 2019.
\newblock \href {https://arxiv.org/abs/1905.07129} {\path{arXiv:1905.07129}}.

\bibitem{su2021cokebert}
Yusheng Su, Xu~Han, Zhengyan Zhang, Yankai Lin, Peng Li, Zhiyuan Liu, Jie Zhou, and Maosong Sun.
\newblock Cokebert: Contextual knowledge selection and embedding towards enhanced pre-trained language models.
\newblock {\em AI Open}, 2:127--134, 2021.
\newblock URL: \url{https://www.sciencedirect.com/science/article/pii/S2666651021000188}, \href {https://doi.org/10.1016/j.aiopen.2021.06.004} {\path{doi:10.1016/j.aiopen.2021.06.004}}.

\bibitem{peters2019knowledge}
Matthew~E. Peters, Mark Neumann, Robert L. Logan~IV au2, Roy Schwartz, Vidur Joshi, Sameer Singh, and Noah~A. Smith.
\newblock Knowledge enhanced contextual word representations, 2019.
\newblock \href {https://arxiv.org/abs/1909.04164} {\path{arXiv:1909.04164}}.

\bibitem{marino2021krisp}
Kenneth Marino, Xinlei Chen, Devi Parikh, Abhinav Gupta, and Marcus Rohrbach.
\newblock Krisp: Integrating implicit and symbolic knowledge for open-domain knowledge-based vqa.
\newblock In {\em Proceedings of the IEEE/CVF Conference on Computer Vision and Pattern Recognition (CVPR)}, pages 14111--14121, June 2021.

\bibitem{li2024knowphish}
Yuexin Li, Chengyu Huang, Shumin Deng, Mei~Lin Lock, Tri Cao, Nay Oo, Bryan Hooi, and Hoon~Wei Lim.
\newblock Knowphish: Large language models meet multimodal knowledge graphs for enhancing reference-based phishing detection, 2024.
\newblock \href {https://arxiv.org/abs/2403.02253} {\path{arXiv:2403.02253}}.

\bibitem{liu2019kbert}
Weijie Liu, Peng Zhou, Zhe Zhao, Zhiruo Wang, Qi~Ju, Haotang Deng, and Ping Wang.
\newblock K-bert: Enabling language representation with knowledge graph, 2019.
\newblock \href {https://arxiv.org/abs/1909.07606} {\path{arXiv:1909.07606}}.

\bibitem{yang2024facts}
Linyao Yang, Hongyang Chen, Zhao Li, Xiao Ding, and Xindong Wu.
\newblock Give us the facts: Enhancing large language models with knowledge graphs for fact-aware language modeling, 2024.
\newblock \href {https://arxiv.org/abs/2306.11489} {\path{arXiv:2306.11489}}.

\bibitem{hou2022enhanced}
Yifan Hou, Guoji Fu, and Mrinmaya Sachan.
\newblock What has been enhanced in my knowledge-enhanced language model?, 2022.
\newblock \href {https://arxiv.org/abs/2202.00964} {\path{arXiv:2202.00964}}.

\end{thebibliography}

\end{document}